\documentclass{article}

\usepackage{acurai_2024}    

\usepackage[utf8]{inputenc} 
\usepackage[T1]{fontenc}    
\usepackage{url}            

\usepackage{booktabs}       
\usepackage{amsfonts}       
\usepackage{nicefrac}       
\usepackage{microtype}      
\usepackage{xcolor}         
\usepackage{subcaption}     
\usepackage{pgf}
\usepackage{pgfplots}       
\pgfplotsset{compat=1.18}
\usepackage{listings}
\usepackage{gensymb}

\usepackage{textcomp} 
\usepackage{siunitx}  

\usepackage[square, numbers]{natbib}
\usepackage[hidelinks]{hyperref} 
\usepackage{geometry}  
\usepackage{amsmath}

\title{100\% Elimination of Hallucinations on RAGTruth for GPT-4 and GPT-3.5 Turbo}

\author{
  Michael C. Wood \\
  Acurai, Inc. \\
  \texttt{michaelw@acur.ai} \\
  \And
  Adam A. Forbes \\
  Acurai, Inc. \\
  \texttt{adamf@acur.ai} \\
}

\begin{document}

\maketitle

\begin{abstract}
The issue of hallucinations in large language models (LLMs) remains a critical barrier to the adoption of AI in enterprise and other high-stakes applications. Despite advancements in retrieval-augmented generation (RAG) systems, current state-of-the-art methods fail to achieve more than 80\% accuracy in generating faithful and factually correct outputs, even when provided with relevant and accurate context. In this work, we introduce Acurai, a novel systematic approach that achieves 100\% hallucination-free responses in LLMs by reformatting queries and context data prior to input. Leveraging a deep understanding of LLM internal representations, the importance of noun-phrase dominance, and the role of discrete functional units (DFUs), Acurai ensures alignment between input context and generated output. We validate this method using the RAGTruth corpus, demonstrating its ability to eliminate 100\% hallucinations for both GPT-4 and GPT-3.5 Turbo. Acurai sets a new standard for achieving consistent, accurate, and faithful AI responses, marking a significant step forward in the development of trustworthy AI systems.

\end{abstract}

\section{Introduction}

One of the largest stumbling blocks to the adoption of LLM-based enterprise chatbots, and more generally "Human-out-of-the-loop-AI", is the problem of hallucinations \cite{rawte2023surveyhallucinationlargefoundation}. Hallucinations in Large Language Models (LLMs) occur when the model generates incorrect or nonsensical information. It has long been believed that hallucinations happen because LLMs rely on patterns in their training data to predict plausible responses, and without domain-specific knowledge or context, they may "guess" answers that sound reasonable but are factually incorrect \cite{10.1145/3688007}. Hallucinations are particularly problematic because these models often present their outputs in an authoritative tone, even when the information is fabricated, leading users to trust incorrect responses \cite{augenstein2023factualitychallengeseralarge}.

For instance, the model might fabricate an answer, such as inventing an academic paper, a non-existent scientific concept, or incorrect historical facts \cite{10.1145/3688007}. To mitigate these issues, strategies like retrieval-augmented generation (RAG) systems, where the model references external verified information, have been proposed as a solution \cite{lewis2021retrievalaugmentedgenerationknowledgeintensivenlp}.

Instead of relying on the model's pre-trained knowledge, RAG utilizes external document collections to provide relevant, up-to-date, and factual information during the response generation process. However, despite more than 5,000 papers on RAG since its introduction in 2020, the promise of RAG remains unfulfilled \cite{gao2024retrievalaugmentedgenerationlargelanguage}.

Although RAG systems can improve accuracy vs. relying on LLM parametric knowledge, \cite{lewis2021retrievalaugmentedgenerationknowledgeintensivenlp}, even state-of-the-art (SOTA) RAG using the latest, cutting-edge LLMs, are still unable to produce more than 80\% accurate results on commonly used RAG QA benchmarks \cite{databricks2024}, \emph{benchmarks that the LLMs are likely trained on}. See Figure 1 for details. As is clearly demonstrated by this chart, the belief that if you send an LLM 100\% factual and relevant data, you will get 100\% factual results, is entirely unfounded.

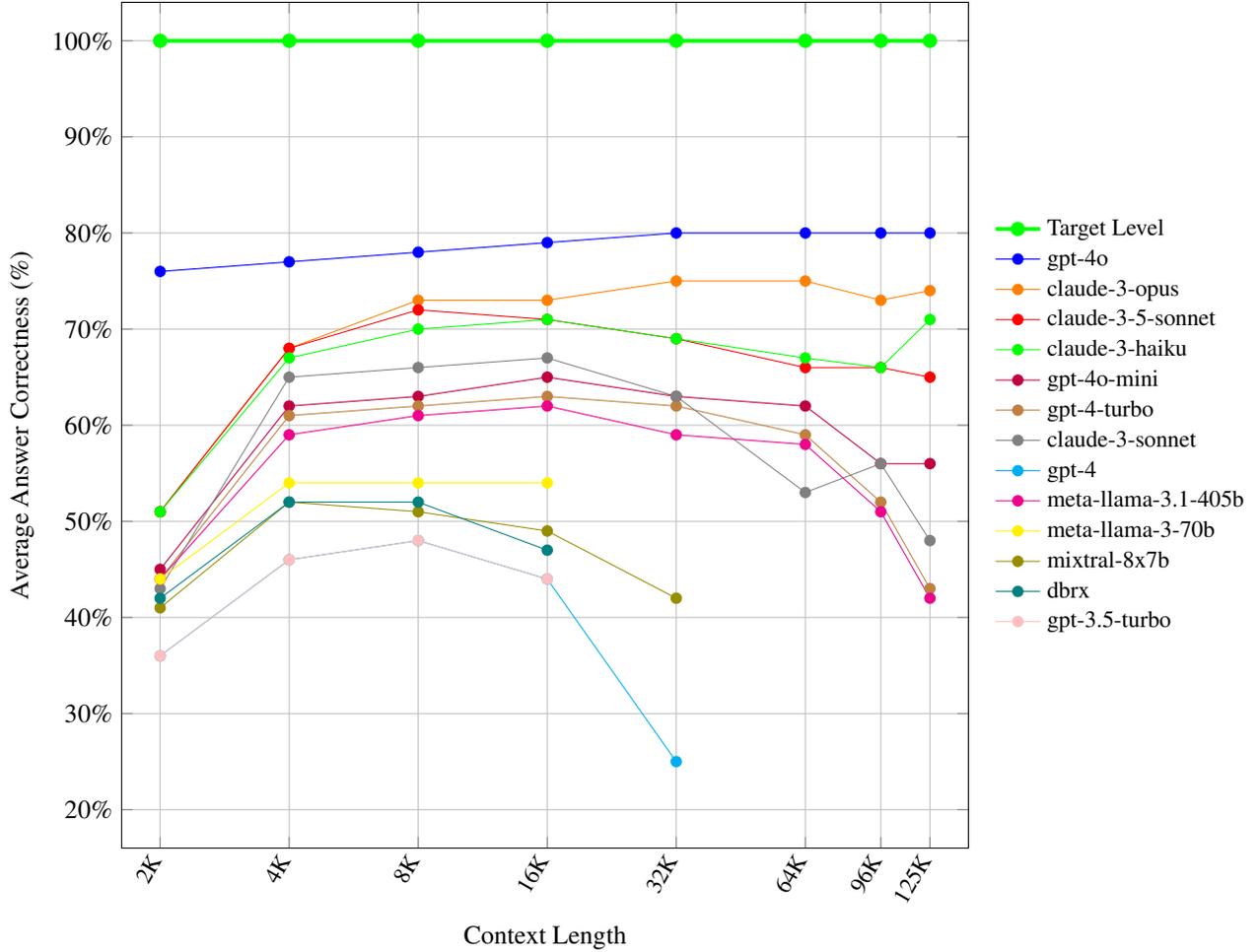
\begin{figure}
\begin{tikzpicture}
\begin{axis}[
    width=13cm,
    height=13cm,
    xlabel={Context Length},
    x tick label style={
        rotate=60,
        anchor=east,
        font=\small
    },
    ylabel={Average Answer Correctness (\%)},
    xmin=2000,
    xmax=125000,
    ymin=0.2,
    ymax=1.0,
    ytick={0.0, 0.1, 0.2, 0.3, 0.4, 0.5, 0.6, 0.7, 0.8, 0.9, 1.0},
    yticklabels={0, 10\%, 20\%, 30\%, 40\%, 50\%, 60\%, 70\%, 80\%, 90\%, 100\%},
    scaled x ticks=false,
    xmode=log,
    log basis x=10,
    enlarge x limits=0.05,
    enlarge y limits=0.05,
    xtick={2000,4000,8000,16000,32000,64000,96000,125000},
    xticklabels={2K,4K,8K,16K,32K,64K,96K,125K},
    grid=major,
    legend style={
        font=\small,
        at={(1.02,0.5)}, 
        anchor=west,
        cells={anchor=west},
        draw=none,
    },
]
\addplot[green, line width=1.5pt, mark=*] coordinates {
    (2000,1.0) (4000,1.0) (8000,1.0) (16000,1.0) (32000,1.0)
    (64000,1.0) (96000,1.0) (125000,1.0)
}; \addlegendentry{Target Level}

\addplot[blue, mark=*, ] coordinates {
    (2000,0.76) (4000,0.77) (8000,0.78) (16000,0.79) (32000,0.80) 
    (64000,0.80) (96000,0.80) (125000,0.80)
}; \addlegendentry{gpt-4o}

\addplot[orange, mark=*] coordinates {
    (2000,0.51) (4000,0.68) (8000,0.73) (16000,0.73) (32000,0.75)
    (64000,0.75) (96000,0.73) (125000,0.74)
}; \addlegendentry{claude-3-opus}

\addplot[red, mark=*] coordinates {
    (2000,0.51) (4000,0.68) (8000,0.72) (16000,0.71) (32000,0.69)
    (64000,0.66) (96000,0.66) (125000,0.65)
}; \addlegendentry{claude-3-5-sonnet}

\addplot[green, mark=*] coordinates {
    (2000,0.51) (4000,0.67) (8000,0.70) (16000,0.71) (32000,0.69)
    (64000,0.67) (96000,0.66) (125000,0.71)
}; \addlegendentry{claude-3-haiku}

\addplot[purple, mark=*] coordinates {
    (2000,0.45) (4000,0.62) (8000,0.63) (16000,0.65) (32000,0.63)
    (64000,0.62) (96000,0.56) (125000,0.56)
}; \addlegendentry{gpt-4o-mini}

\addplot[brown, mark=*] coordinates {
    (2000,0.44) (4000,0.61) (8000,0.62) (16000,0.63) (32000,0.62)
    (64000,0.59) (96000,0.52) (125000,0.43)
}; \addlegendentry{gpt-4-turbo}

\addplot[gray, mark=*] coordinates {
    (2000,0.43) (4000,0.65) (8000,0.66) (16000,0.67) (32000,0.63)
    (64000,0.53) (96000,0.56) (125000,0.48)
}; \addlegendentry{claude-3-sonnet}

\addplot[cyan, mark=*] coordinates {
    (2000,0.36) (4000,0.46) (8000,0.48) (16000,0.44) (32000,0.25)
}; \addlegendentry{gpt-4}

\addplot[magenta, mark=*] coordinates {
    (2000,0.44) (4000,0.59) (8000,0.61) (16000,0.62) (32000,0.59)
    (64000,0.58) (96000,0.51) (125000,0.42)
}; \addlegendentry{meta-llama-3.1-405b}

\addplot[yellow, mark=*] coordinates {
    (2000,0.44) (4000,0.54) (8000,0.54) (16000,0.54)
}; \addlegendentry{meta-llama-3-70b}

\addplot[olive, mark=*] coordinates {
    (2000,0.41) (4000,0.52) (8000,0.51) (16000,0.49) (32000,0.42)
}; \addlegendentry{mixtral-8x7b}

\addplot[teal, mark=*] coordinates {
    (2000,0.42) (4000,0.52) (8000,0.52) (16000,0.47)
}; \addlegendentry{dbrx}

\addplot[pink, mark=*] coordinates {
    (2000,0.36) (4000,0.46) (8000,0.48) (16000,0.44)
}; \addlegendentry{gpt-3.5-turbo}

\end{axis}
\end{tikzpicture}
\caption{Long Context RAG Performance of LLMs \cite{databricks2024}. \emph{Note that best case, fully 1 in 5 answers are incorrect.}}
\label{fig:model-performance}
\end{figure}

Thus, finding a systematic method to enable LLMs to generate hallucination-free, accurate responses \textbf{even when the LLM is provided clearly stated facts} was an open problem in the field.

In this work, we explore a systematic approach to achieving 100\% accurate, hallucination-free LLM output by modifying both the query and the context data prior to them being sent to the LLM.

\section{Related Works}

\paragraph{RAG Hallucination Evaluation Benchmarks}

Many benchmarks are designed to identify hallucinations in the context of parametric knowledge \cite{lin2022truthfulqa}\cite{wei2024simpleqa}.  Others have been developed to quantify the level of hallucination generated by RAG pipelines \cite{friel2024ragbench} (e.g. a Retriever plus a Generator, typically an LLM such as ChatGPT-4). We chose to use the RAGTruth corpus, a third-party dataset that documents RAG-based hallucinations for GPT-4, GPT-3.5 Turbo, and other popular LLMs \cite{niu2024ragtruth}. The dataset provides three relevant passages for each query. The same query and passages were sent to a variety of LLMs. The dataset lists, among other things, Evident Conflicts and Subtle Conflicts seen in the responses of each model. In other words, the dataset lists the query/passages combinations that cause hallucinations in each of the studied LLMs. 

\paragraph{Internal Representation of LLMs}

A key to our method for eliminating hallucinations relates to how LLMs self-organize during training. Acurai's hallucination elimination model states that LLMs self-organize around noun phrases. OpenAI and Anthropic's recent studies into how their LLMs work precisely aligns with our own model of LLM self-organization. Specifically, OpenAI and Anthropic both refer to "features" \cite{gao2024openai} and "interpretable features" \cite{anthropic2024}, and both identified a handful of "features" in their neural networks that are important to producing any given output. These features are akin to the small set of concepts a person might have in mind when reasoning about a situation. We refer to such features as Noun-Phrases, and our Noun-Phrase Dominance Model that predates the OpenAI and Anthropic studies posited that LLMs self-organize around such noun phrases during training \cite{mcw2024}.

\paragraph{Long Context RAG}

Recently, LLMs have greatly increased their context window, allowing a very large amount of text to be sent along with the prompt. Before the advent of long-context language models, RAG was a key solution for overcoming the limitations of small context window sizes. Some researchers have examined if a sufficiently large context window could effectively replace RAG altogether. Databricks examined the effects of increased context size on various modern LLMs. The net result was that no LLM, regardless of the amount of data sent to it, was able to exceed greater than 80\% accuracy on various QA benchmarks \cite{databricks2024}.

\paragraph{Faithfulness and Correctness}

Two terms that are important to understand as it relates to evaluating RAG and LLMs are Faithfulness and Correctness. Faithfulness evaluates whether the generated output accurately reflects the information contained in the retrieved documents \cite{adlakha2024faithful}. A response is considered faithful if it does not introduce information that is absent from the retrieved sources, and adheres closely to the input data. This criterion measures how consistently the LLM response aligns with the retrieved content. A perfect faithfulness score means there is no "hallucination" caused by deviation from the content (such as by introducing new facts or interpretations). In other words, a faithful response directly reflects the content of the retrieved documents, with all statements fully supported by the retrieved data. 

Correctness, on the other hand, assesses the factual accuracy of the output within a broader context \cite{adlakha2024faithful}. A response is deemed correct if it aligns with established facts, even when those facts are not explicitly present in the retrieved documents. Correctness measures how well the generated answer matches verified external knowledge. For instance, a correct response is factually accurate based on real-world information, even if some of that information originates from the model's parametric knowledge rather than the retrieved (non-parametric) sources.

Faithfulness ensures that the response stays true to the retrieved information, while correctness ensures that the response is factually accurate, regardless of its source.

\section{Explanation of Acurai}

\paragraph{Terms}

We use the terms "accurate" and "faithful" herein interchangeably. For our primary use case (enterprise chatbots), the goal is to provide answers that are faithful to the information provided. For example, if an enterprise customer is a car company, they want their chatbot to be faithful to the documents provided to the chatbot, which may say their cars are "the best". This faithful answer may not be an objectively quantifiable answer; and perhaps even according to a third party, this company's cars are not, in fact, "the best". In many chatbots, correctness can be ambiguous or debatable, depending on the topic and question asked ("Who is the greatest opera singer of all time?"), whereas faithfulness can always be measured in concrete terms.

In a RAG-based chatbot, we refer to hallucinations as referring to any deviation from the provided context. Importantly, \textbf{LLMs can still hallucinate even when clearly written facts are sent along with the query}. For example, ChatGPT-3.5 Turbo was provided the following clearly written statements about \emph{calcium}: "Calcium is a sliver-grey metal. Calcium melts at 840\degree C. Calcium boils at 1484\degree C to produce monatomic gas. ..." Remarkably, the LLM stated that all these properties belonged to \emph{magnesium} when given the following instruction: "Extract all facts about magnesium from the following passages." \cite{wood2024youtubea}

The calcium statements could not be more clearly written. The prompt also clearly asks about magnesium. Nevertheless, the LLM treated \emph{magnesium} as if it is the same thing as \emph{calcium}. Acurai's Noun-Phrase Dominance Model says that all hallucinations occur when the LLM mistakes two distinct Noun-Phrases as being the same thing. More specifically, the LLM does so when the distinct noun phrases are \emph{semantically similar}, such as is the case with calcium and magnesium \cite{wood2024youtubeb}.

The corollary is that RAG-based hallucinations can be eliminated by ensuring that the LLM is never sent semantically similar noun phrases that refer to distinctly different things. By way of example, Acurai takes the following steps, among others, to accomplish this goal.

Let's examine the following query that generated a hallucination in the RAGTruth Corpus: 

\begin{lstlisting}
"What are the chemical and physical properties of calcium and magnesium?"
\end{lstlisting}

\subsection{Step One: Split the query to separate noun phrase collisions}

Noun-Phrase collisions can occur when semantically similar terms refer to discretely different noun phrases. For example, 'car' and 'automobile' are not noun-phrase collisions because they can refer to the same object even though they are semantically similar. However, calcium and magnesium can cause noun-phrase collisions because they are semantically similar (i.e. their vector embeddings have a high cosine similarity score) even though they refer to discretely separate things.

Thus, prompts need to be rewritten to split noun-phrase colliding references into separate queries. In the RAGTruth example, there are two noun-phrase collision pairs: chemical properties \& physical properties, magnesium \& calcium. Therefore, the prompt must be split into at least four queries, such as the following:

\begin{lstlisting}
What are the chemical properties of magnesium?
What are the physical properties of magnesium?
What are the chemical properties of calcium?
What are the physical properties of calcium?
\end{lstlisting}

These queries are now clear and distinct, each focusing on either the chemical or physical properties of magnesium or calcium. Most importantly, the queries are devoid of noun-phrase collisions.

\subsection{Step Two: Only send simple statements devoid of noun phrase collisions to the LLM}

In the RAGTruth study, the following three passages were sent to the LLM along with the query:

\begin{lstlisting}
"The chemical properties of calcium are reacts with oxygen and reacts with water. There are other chemical properties, but not all of them are true for calcium.These are the two that I know.he chemical properties of calcium are reacts with oxygen and reacts with water. There are other chemical properties, but not all of them are true for calcium."

"Chemical and Physical Properties of Magnesium. Magnesium is one of the most important elements that is present in many compounds as well as alloys. It is widely used as a chemical reagent, desulfurization agent, and vital ingredient in fireworks.It finds multiple applications due to its unique chemical and physical properties.s mentioned in the chemical properties, magnesium is also present in many other compounds like dolomite, magnesium carbonate (that is also known as magnesite), and magnesium sulfate (which is also known by the name epsomite)."

"Calcium: Physical Properties --silver-grey metal. melts at 840\u00b0C, boils at 1484\u00b0C to produce monatomic gas. density 1540 kg/m\^3. conductor of electricity but a poor one com \u2026pared to most other metals.Diamagnetic. Chemical properties --tarnishes rapidly in air to produce a powdery, flaky oxide coating.Reacts steadily with water, giving off bubbles of hydrogen and a solution/slurry or alkaline, sparingly soluble calcium hydroxide. Flammable at high temperatures with oxygen, or air.an also burn in nitrogen to form calcium nitride or carbon dioxide to form calcium carbonate. Nearly all compounds are in oxidation state +2, and the water chemistry of calcium is dominated by the hydrated Ca(2+) ion"
\end{lstlisting}

Acurai rewrites the passages, transforming them into Fully-Formatted Facts (FFFs). FFFs are simple, self-contained statements that are devoid of noun-phrase collisions. To achieve the latter, the simple, self-contained statements are paired with their respective queries. In other words, only statements regarding magnesium are sent along with the magnesium queries, and only statements about calcium are sent along with the calcium queries.

For example, Acurai sent the following fact sets along with the two calcium queries:

\begin{lstlisting}
Query: 
    what are some physical properties of calcium
Fact Set: 
    Section 1:
        Calcium is a silver-grey metal.
        Calcium melts at 840°C.
        Calcium boils at 1484°C to produce monatomic gas.
        Calcium's density is 1540 kg/m^3.
        Calcium is diamagnetic.

Query: 
    what are some chemical properties of calcium
Fact Set: 
    Section 1:
        The chemical properties of calcium react with oxygen.
        The chemical properties of calcium react with water.
    Section 2:
        Calcium melts at 840°C.
        Calcium boils at 1484°C to produce monatomic gas.
        Calcium's density is 1540 kg/m^3.
        Calcium is a conductor of electricity.
        Calcium is diamagnetic.
        Calcium reacts steadily with water.
        Calcium gives off bubbles of hydrogen.
        Calcium produces a solution/slurry of alkaline, sparingly soluble calcium hydroxide.
        Calcium is flammable at high temperatures with oxygen.
        Calcium is flammable at high temperatures with air.
        Calcium can also burn in nitrogen to form calcium nitride.
        Calcium can also burn in carbon dioxide to form calcium carbonate.
        Nearly all Calcium's compounds are in oxidation state +2.
        The water chemistry of calcium is dominated by the hydrated Ca(2+) ion.
\end{lstlisting}

\subsection{Step Three: Remap any modified text}

Sometimes, passages need to be rewritten to remove noun-phrase collisions. For example, a statement about "cruise control" might be found within a passage regarding Cruise LLC. In this case, the word "cruise" is used to refer to two different things, causing a potential noun-phrase collision. This can be remedied by replacing "Cruise LLC" with a placeholder name. Then the name "Cruise LLC" is mapped back into the response anywhere the placeholder is used.

Such mapping also is used for items that inherently contain noun-phrase collisions within themselves, such as references, citations, part numbers, PubMed IDs, etc. Each individual reference can have self-contained noun-phrase collisions, which is why they cause an exceedingly high rate of hallucinations in even the most capable of LLMs \cite{wood2024youtube}.  For example, even ChatGPT hallucinates 93\% of the time when providing PubMed IDs \cite{bhattacharyya2023}. Moreover, ChatGPT hallucinates 64\% of the time for volume numbers, 64\% for page numbers, and 60\% for year of publication \cite{bhattacharyya2023}.

Acurai's solution is to replace all such references with single-token placeholders. Then remap the original references in the response where each placeholder is used.

\section{Experiments}

\subsection{RAGTruth Dataset Caveats}

When assessing Acurai's faithfulness, we solely compared the final response to the provided passages, without considering the associated annotator's notes. This is due to three defects in the RAGTruth corpus. 

\paragraph{Missing Instructions}

Some passages contain missing instructions. For example, the query "How do I reheat chicken already cooked?" (Response ID: 8285) is sent with one of the following passages:

\begin{quote}
Step 1. Place the chicken on a microwave-safe plate or dish. If you are reheating pieces, place the largest, meatiest pieces towards the outside of the dish and smaller pieces in the center. Food on the outer edge of the dish cooks faster.tep 4. Reheat the chicken for two to three minutes, then turn the pieces over and stir the sauce. Resume cooking for an additional two to three minutes, or until the center of the thickest piece of chicken is cooked to an internal temperature of at least 165 degrees Fahrenheit.
\end{quote}

Notice that there is no step 2 or step 3 provided. The passage jumps from step 1 to "tep 4" (sic). 

\paragraph{Wrong information}

Some queries are sent with incorrect information. For example, query "history of minimum wage" (Response ID: 9824) is sent with one of the following passages:

\begin{quote}
No state minimum wage law. The minimum wage in the United States is a network of federal, state, and local laws. Employers generally must pay workers the highest minimum wage prescribed by federal, state, or summer local law. As of July 2016, the federal government mandates a nationwide minimum wage of \$7.25 per hour. As of October 2016, there are 29 states with a minimum wage higher than the federal minimum.
\end{quote}

Notice that there is conflicting information. The first sentence says there is no state minimum wage law, while the last sentence states that there are 29 states with a minimum wage higher than the federal minimum wage.

\paragraph{Wrong annotations} 

RAGTruth contains incorrect annotations. For example, the query "what did the industrial revolution do for society" (Response ID: 9692) was mistakenly labeled as an "EVIDENT CONFLICT". The original text was "occurred between the late 1700s and the early 1900s". The LLM re-wrote that as "occurred between the late 18th and early 20th century". Late 18th century is the same as late 1700s, and early 20th century is the same as early 1900s. The annotator wrongly considered these to be evident conflicts.

\subsection{Evaluation}

Therefore, we independently compared Acurai's response to the provided passages to determine if the responses deviated from the provided passages in any manner whatsoever. The Acurai responses are listed in the following Github repo for independent review and validation: \url{https://github.com/AcuChat/acurai-RAGTruth-conflict-resolution}.

In order to evaluate the accuracy of our system, we chose to examine Acurai's faithfulness on the following four datasets in the RAGTruth Corpus:

\begin{itemize}
\item GPT-3.5 Turbo Subtle Conflict
\item GPT-3.5 Turbo Evident Conflict
\item GPT-4 Subtle Conflict
\item GPT-4 Evident Conflict
\end{itemize}

A copy of the datasets is provided in the accompanying Github repo: \url{https://github.com/AcuChat/acurai-RAGTruth-conflict-resolution}.

Every response in each of the datasets is a hallucination. To measure Acurai, the same passages and queries were sent to the same model with Acurai transforming the model's inputs and outputs. In other words, for GPT-3.5 Turbo Evident Conflicts, the identical query and passages were sent to the same GPT-3.5 Turbo model used in the RAGTruth study. However, the inputs and outputs of this model were transformed by Acurai in accordance with the example steps listed above.

The resulting Acurai response was then compared to the original passages. If the Acurai response deviated from the passages, it would be considered to be a hallucination. If the entire response match the statements in the passage, then it would be considered hallucination-free.

The objective was to measure if Acurai could transform each of the hallucinating models into accurate ones. As documented in the Github, Acurai eliminated 100\% of the hallucinations in all four data sets. See Figure 2 for results.

\begin{figure}[htbp]
    \centering
\begin{tikzpicture}
    \begin{axis}[
        xbar,
        width=9cm,
        height=6cm,
        bar width=0.6cm,
        enlarge y limits=0.3,
        axis lines*=left,
        xlabel={Percentage Accurate \& Hallucination-Free},
        symbolic y coords={{GPT-4 Evident Conflict}, {GPT-4 Subtle Conflict}, {GPT-3.5 Turbo Evident Conflict}, {GPT-3.5 Turbo Subtle Conflict}},
        ytick=data,
        nodes near coords={\pgfmathprintnumber\pgfplotspointmeta\%},  
        nodes near coords align={horizontal},
        xtick={0,20,40,60,80,100},
        xticklabel={\pgfmathprintnumber\tick\%},  
        xmin=0, xmax=100,
    ]
    
    \addplot[fill=green!50] coordinates {
        (100,{GPT-3.5 Turbo Subtle Conflict}) 
        (100,{GPT-3.5 Turbo Evident Conflict}) 
        (100,{GPT-4 Subtle Conflict}) 
        (100,{GPT-4 Evident Conflict})
    };
    \end{axis}
\end{tikzpicture}
\caption{Acurai RAGTruth Results by Model}
\end{figure}
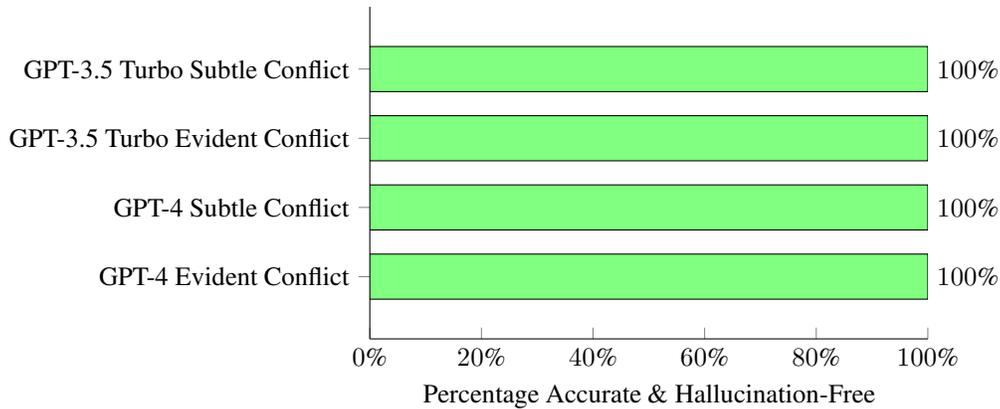

In summary, Acurai achieved a perfect score, giving faithful, hallucination-free answers to all questions in the categories of Subtle Conflict and Evident Conflict for both GPT-3.5 Turbo and ChatGPT-4. 

Details of methodology:

\begin{enumerate}
\item Same LLM and Model As RAGTruth
\item Same Temperature As RAGTruth
\item Same Query As RAGTruth
\item Same Passages As RAGTruth
\item Query \& Passages systematically rewritten per Acurai rules
\item Rewritten Query \& Passages then sent to identical LLM and Model
\item Compared the outputs to the original passages
\end{enumerate}

All of the questions and responses from GPT-3.5 Turbo, GPT-4 and Acurai are included in the accompanying GitHub repo, for independent verification. For a sample full question and response, see Appendix A: Sample RAGTruth Question.

\subsection{Confidence Interval}

Due to the 100\% success rate, the confidence interval was calculated using the Wilson Score Interval.

\[
\hat{p} = \frac{X + z^2/2}{n + z^2}
\]

\[
\text{Confidence Interval} = \hat{p} \pm z \sqrt{\frac{\hat{p}(1 - \hat{p})}{n + z^2}}
\]

where:
\begin{itemize}
\item $z$ is the z-value for a 95
\item $p$ is the observed proportion of successes (1.0),
\item $n$ is the sample size (37).
\item $X$ is the number of successes (37).
\end{itemize}

This results in the following proportion: 95\% CI [0.91, 1]. In other words, \textbf{Acurai is expected to eliminate between 91\% to 100\% of hallucinations in the general population} under the same conditions. The unprecedented nature of this result underscores Acurai's potential to resolve the issue of hallucinations in ChatGPT-4 and ChatGPT-3.5 Turbo.

\section{Limitations}

The RAGTruth dataset contains factually correct passages that are sent to the Large Language Model (LLM). Thus, the hallucination elimination solely applies to RAG-based implementations that only send factual chunks.
   
The RAGTruth dataset provides three passages. Thus, Acurai's effectiveness was solely measured when a relatively small number of passages are sent along with the query. In contrast, modern RAG implementations like OP-RAG \cite{yu2024oprag} and Long RAG \cite{jiang2024longrag} typically send hundreds of passages to the LLM. Thus, additional work needs to be done to measure Acurai's effectiveness in situations where a large number of irrevelant passages are sent. And/or additional work needs to be done to implement a RAG-based chatbot that solely sends relevant passages to Acurai.
   
The RAGTruth passages are almost exclusively relevant to the queries. Thus, a similarly performing RAG-based chatbot would need to be implemented to mirror the hallucination-free results.

The unprecedented confidence interval of 91\% to 100\% was achieved on one specific family of LLMs. Additional work needs to be done to determine if Acurai resolves the issue of hallucinations for other model families such as Llama, Gemini, and Granite.
   
Implementing this system results in higher operational costs compared to using the base LLM alone, due to the additional computational resources required for processing. This limitation can be mitigated by pairing Acurai with cheaper, open-source LLMs such as Llama 3.1 405B.
   
The system also introduces additional latency to query responses, as it must pre-process both the query and passages before submitting to the LLM, leading to slower response times compared to direct LLM queries. This limitation could be mitigated by converting the document information into Fully-Formatted Facts prior to storage, removing the burden from the LLM at the time of query.

\section{Conclusions}

The output of large language models (LLMs) is known to be influenced by variations in queries and passages. However, Acurai shows that systematically altering them can result in consistent accuracy. This conclusion is supported by the fact that all four datasets achieved 100\% accuracy using the same transformation method. Each Question / Passages pair underwent an identical process, converting hallucinations into fully accurate responses.

Importantly, Acurai systematically prevents hallucinations from occurring; as opposed to blocking hallucinations after they already exist. This means that users receive accurate responses, instead of being informed that a response cannot be given. Prioritizing the prevention of hallucinations is more effective than focusing on blocking them.

The dataset sample size establishes a confidence interval between 91\% to 100\% (with a 95\% confidence level). No other system or method has been shown to systematically eliminate hallucinations across any sample size. Thus, our research introduces a transformative framework that redefines the limits of what is achievable in hallucination elimination.


\newpage
\appendix

\section{Appendix: Sample RAGTruth Question \& Answer}

Example Question: "benefits of ice for neck"

Model: GPT-4-0613

Response ID: 7969

\rule{\textwidth}{0.5pt}

\textbf{GPT-4-0613 response without Acurai}

Applying ice to the base of the neck has been suggested to provide a variety of benefits. This includes improved digestion and sleep, reduced thyroid issues and PMS symptoms, alleviation of common colds, headaches or toothaches, and reduced risks associated with lung and cardiovascular diseases. It can also potentially boost mood and mental health.

\rule{\textwidth}{0.5pt}

\textbf{Acurai response using GPT-4-0613}

\begin{center}
\textbf{Benefits of ice for neck}
\end{center}

Placing an ice cube at the base of your skull, on your neck, can provide a variety of benefits. The technique is simple: just let the ice cube rest there for about 20 minutes. Besides providing an overall boost to your mood and mental health, consistent use of this technique for a period of one month can lead to potential health benefits such as improved digestion and sleep. Additionally, it's been found to help reduce symptoms associated with thyroid issues and Pre-menstrual Syndrome (PMS). But the benefits don't stop there, this technique can also help to cure common colds and alleviate discomfort from headaches or toothaches. Furthermore, using this technique can contribute to a reduction in risks associated with lung and cardiovascular diseases.

\begin{center}
\textbf{Specifics}
\end{center}

\paragraph{Detail 1:}

All it requires is that every day you lay on your tummy. Place an ice cube at the base of your skull on your neck. See the point on the neck shown in the video. Allow it to rest there for 20 minutes. Do this in the morning on an empty stomach. Do this before you go to sleep.

\rule{\textwidth}{0.5pt}

\textbf{Passages sent along with query:}

All it requires is that every day you lay on your tummy. Place an ice cube at the base of your skull on your neck (see the point on the neck shown in the video), and allow it to rest there for 20 minutes. Do this in the morning on an empty stomach and before you go to sleep. It has been said that this technique can provide a variety of benefits to your body, as well as boost your mood and mental health. After consistent use for a month potential health benefits include improved digestion and sleep, reduced thyroid issues and PMS symptoms, cure common colds, alleviate headaches or toothaches, and reduce overall risks associated with lung and cardiovascular diseases.

For the hardcore: Ice bath Grab three bags of ice from a convenience store and fill your bath tub halfway full with cold water. Pour the ice in. (The first few times you take an ice bath, only immerse your lower body, from the hips down. After you get more comfortable with sitting in the ice bath, begin slowly lowering your upper torso until submerged, up to your neck if you can handle it.) Sip a cup of hot tea and read a magazine to take your mind off the bath.

Woman Places An Ice Cube On This Spot Of Her Neck For A Month. I Had No Idea It Would Do THIS For as long as man has existed on earth he has been searching for a fountain of youth.

\rule{\textwidth}{0.5pt}

\textbf{Error with GPT-4-0613 response without Acurai:}

SUBTLE CONFLICT

Original: base of your skull on your neck

Generated: base of the neck

\rule{\textwidth}{0.5pt}

\textbf{Error with Acurai response using GPT-4-0613:}

NONE

\rule{\textwidth}{0.5pt}

\end{document}